%% file: main.tex
\documentclass[10pt,twocolumn,letterpaper]{article}
\usepackage[noend,linesnumbered,ruled,vlined]{algorithm2e} 
\usepackage[pagenumbers]{cvpr} 


\usepackage{graphicx}
\usepackage{tabularx}
\usepackage{multirow}
\usepackage{tikz}
\usepackage{comment}
\usepackage{colortbl}
\usepackage[accsupp]{axessibility}  
\usepackage{xcolor} 
\usepackage[export]{adjustbox}
\usepackage{graphicx}
\usepackage{multirow}
\usepackage{tikz}
\usepackage{comment}
\usepackage{colortbl}
\usepackage{xcolor} 
\usepackage[export]{adjustbox}
\definecolor{cvprblue}{rgb}{0.21,0.49,0.74}
\usepackage[pagebackref,breaklinks,colorlinks,allcolors=cvprblue]{hyperref}

\def\paperID{xxxxx} 
\def\confName{CVPR}
\def\confYear{2025}

\title{Foveated Instance Segmentation}

\author{
Hongyi Zeng\textsuperscript{$\dag$} \enspace Wenxuan Liu\textsuperscript{$\dag$}  \enspace  Tianhua Xia\\
New York University\\
\and
Jinhui Chen \enspace Ziyun Li\\
Meta Reality Labs\\
\and
Sai Qian Zhang\\
New York University\\
}

\begin{document}
\maketitle
\let\thefootnote\relax\footnotetext{\^\dag \textup{Equal Contribution}}
\begin{abstract}
Instance segmentation is essential for augmented reality and virtual reality (AR/VR) as it enables precise object recognition and interaction, enhancing the integration of virtual and real-world elements for an immersive experience. However, the high computational overhead of segmentation limits its application on resource-constrained AR/VR devices, causing large processing latency and degrading user experience. 
In contrast to conventional scenarios, AR/VR users typically focus on only a few regions within their field of view before shifting perspective, allowing segmentation to be concentrated on gaze-specific areas. This insight drives the need for efficient segmentation methods that prioritize processing instance of interest, reducing computational load and enhancing real-time performance.

In this paper, we present a~\textit{foveated instance segmentation} (FovealSeg) framework that leverages real-time user gaze data to perform instance segmentation exclusively on instance of interest, resulting in substantial computational savings. Evaluation results show that FSNet achieves an IoU of 0.56 on ADE20K and 0.54 on LVIS, notably outperforming the baseline. The code is available at \url{https://github.com/SAI-Lab-NYU/Foveated-Instance-Segmentation}
\end{abstract}

\section{Introduction}
\label{sec:intro}

Semantic segmentation~\cite{long2015fully, badrinarayanan2017segnet,kirillov2023segment, xie2021segformer}, a fundamental task in computer vision, involves partitioning an image into meaningful regions to facilitate the analysis and interpretation of its visual content. Instance segmentation~\cite{he2017mask, bolya2019yolact,wang2020solov2,yang2019video} takes this further by identifying and delineating each individual object instance within an image, which plays a critical role in augmented reality (AR) as it enables precise object recognition and separation in real-world scenes, allowing for more accurate interaction, manipulation, and integration of virtual elements with physical objects for an immersive and context-aware user experience~\footnote{Experiments were conducted at New York University (NYU).}. 
\begin{figure}[t]
\centering
\includegraphics[width=\linewidth]{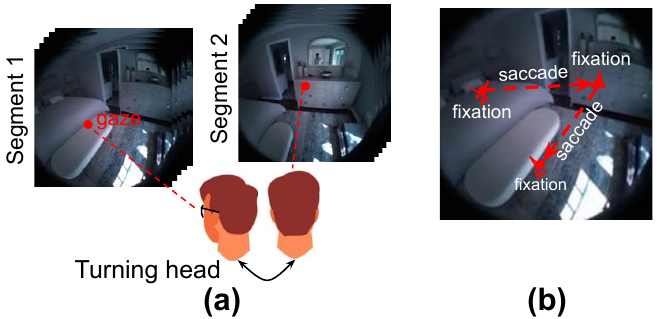}
\caption{(a) An example on gaze location for the AR user. (b) Trace of eye gaze within a segment.}
\label{fig:ar_motivation}
\end{figure}

Despite its importance, the segmentation task poses substantial computational challenges, particularly on resource-limited AR/VR devices, largely due to the high resolution of input images captured by these devices. For instance, the Meta Ray-Ban glasses feature a 12-megapixel camera capable of recording 1440P video~\cite{rayban}, which results in significant computational overhead during instance segmentation. This high data volume results in considerable computational latency, which can severely limits performance and responsiveness, ultimately degrading the overall user experience by impeding real-time interaction and fluidity.

In contrast to conventional use cases, AR/VR device users have a unique behavior: they tend to focus on only some small areas within a view before shifting to another view. For instance, as shown in Figure~\ref{fig:ar_motivation}, a user wearing AR glasses stands in a bedroom. In the left part of Figure~\ref{fig:ar_motivation} (a), the user looks at the bed for a few seconds before turning head to look at the wardrobe, as depicted in the right part of Figure~\ref{fig:ar_motivation} (a). In this scenario, the sequential video frames can be divided into two segments based on head movement. Within the first segment, where the frames are highly similar, the gaze is primarily focused on the bed, allowing instance segmentation to be performed only on the bed. Similarly, in the second segment, segmentation can be limited to the wardrobe.
This insight offers an inherently efficient solution for instance segmentation in AR/VR environments by prioritizing the processing of instances of interest (IOI) as determined by the user’s gaze. By focusing computational resources on these targeted areas, it is possible to significantly reduce processing workload and computational costs, enhancing the real-time performance of AR/VR applications and improving the overall user experience. This aligns naturally with the paradigm of~\textit{foveated rendering}~\cite{patney2016towards}, which enhances graphical performance by rendering images at full resolution only in the area where the user’s gaze is directed, while reducing detail in the peripheral vision to conserve computational resources.

\begin{figure}[t]
\centering
\includegraphics[width=0.95\linewidth]{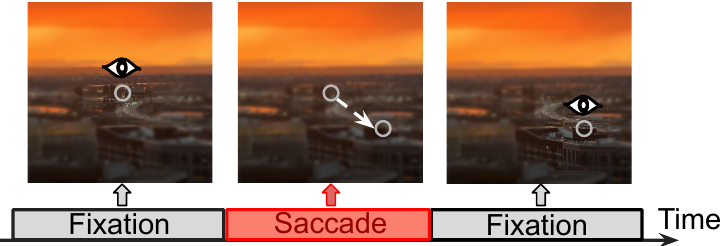}
\caption{An example on fixation and saccade of human eye.}
\label{fig:eye_example}
\end{figure}
In this paper, we propose a novel approach to instance segmentation, termed~\textit{foveated instance segmentation}, by adopting a foveated processing strategy, where segmentation is applied solely at the instance where human gaze locates, eliminating the need to process the entire image. While this approach holds significant potential for efficiency, it also presents several challenges. The first challenge is designing a deep neural network (DNN) framework capable of processing only the IOI associated with the gaze location. The second challenge involves leveraging the temporal dynamics of human gaze to further reduce redundant computations and enhance processing efficiency.  Our contribution can be summarized as follow:
\begin{itemize}
\item We propose a novel and crucial insight into instance segmentation that leverages human eye behavior to reduce computational costs in AR/VR environments.   
\item We introduce~\textit{FSNet}, a plug-and-play instance segmentation neural network that takes a high-resolution input image and gaze location, efficiently performing instance segmentation solely on the instance of interest (IOI). FSNet can integrate with any existing segmentation network, significantly enhancing its generalizability.
\item Building on FSNet, we further introduce \textit{FovealSeg}, an efficient foveated instance segmentation framework designed for real-time AR/VR processing. FovealSeg leverages temporal similarity between consecutive frames and human gaze patterns to optimize segmentation, enhancing performance for dynamic AR/VR environments.

\end{itemize}

\section{Background and Related Work}
\label{sec:bg}
 \subsection{Human Eye Behavior}
\label{sec:bg:human-eye}

The human eye functions in three primary modes of movement, each with distinct roles:~\textit{fixation}, when the eye is still and focused on a single point; ~\textit{saccadic movements}, rapid, jerky movements that shift the gaze from one target to another; and~\textit{smooth pursuit}, when the eye smoothly follows a moving object. Among these, smooth pursuit is less common, while fixation and saccadic movements dominate most of our visual activity, as shown in Figure~\ref{fig:eye_example}. 
During the fixation, human gaze remains focused around a single point, with visual acuity mostly concentrates at the region around the gaze location, and drops greatly outside the region, leading to decreased perception in peripheral vision~\cite{wenxuan_2025}.
\begin{figure}[t]
\centering
\includegraphics[width=0.9\linewidth]{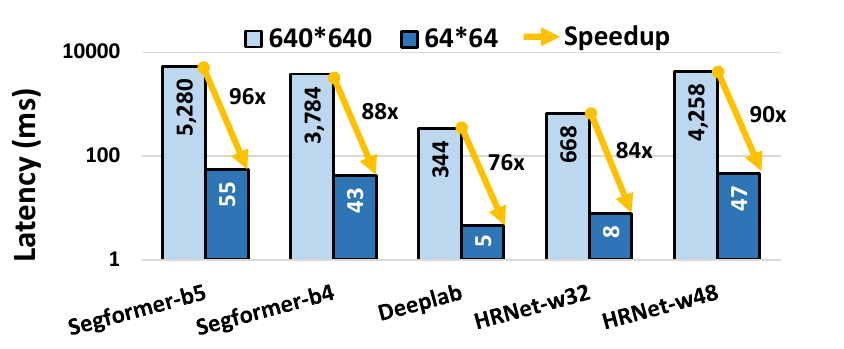}
\caption{Processing latency of segmentation task on edge GPUs.}
\label{fig:hardware_motivation}
\end{figure}

Additionally, humans make one to three saccadic eye movements per second~\cite{kowler2011eye, fabius2019time, kwak2024saccade}, with each saccade duration typically ranges from 20-200ms and speed reaching speeds over 200° per second~\cite{robinson1964mechanics}. Human saccade are essential for scanning the environment efficiently. During a saccade, visual information becomes momentarily blurred due to the high speed of the eye movement, a phenomenon known as saccadic blur~\cite{matin1974saccadic, campbell1978saccadic,idrees2020perceptual}. Once the eye reaches its new target, visual clarity is restored, and the brain integrates information from both the fixation and motion periods to create a stable perception of the environment. Previous studies have shown that perceptual saccadic suppression reaches its peak at the maximum velocity of saccadic movement, leading to a substantial reduction in detectability by at least $75\%$~\cite{idrees2020perceptual}. This allows visual operations to be temporarily halted during the saccade stage without affecting the user's visual experience, as demonstrated in~\cite{Kim2019foveatedar, Loschky2007lateupdate}.

Figure~\ref{fig:ar_motivation} (b) shows an example on trace of the human gaze  within a single frame segment. Initially, the gaze is fixed on the bed, followed by a saccade where it rapidly shifts to the wardrobe. During the fixation stage, minor movements may occur due to natural eye muscle activity. In this context, only two instances need to be detected (i.e., the bed and the wardrobe), and processing during the saccade phase is unnecessary, as human vision is less sensitive and the gaze moves rapidly during saccades. 
Previous studies~\cite{Arab2017saccade,Arab2023saccade} have shown that saccades can be reliably detected by analyzing gaze location changes over a unit time period. When this change surpasses a certain high threshold, a saccade is identified. A similar criterion can be applied to detect the end of a saccade.

\subsection{Instance Segmentation Cost in AR/VR device}
\label{sec:bg:hardware_evaluation}

Since modern AR/VR devices do not allow users to modify the internal algorithm, we use the GPU simulation tool GPGPU-Sim~\cite{khairy2020accelsim} to simulate hardware performance. We configure GPGPU-Sim to model the Jetson Orin NX~\cite{nvidia-gpu}, a widely used edge GPU in AR/VR environments~\cite{zhang2024boxr, zhang2024co, pancrisp, he2024omnih2o, gilles2023holographic}. We simulate the hardware performance by measuring the processing latency of four popular deep neural network (DNN) architectures for segmentation: Segformer~\cite{xie2021segformer}, Deeplab~\cite{chen2017deeplab}, and HRNet~\cite{wang2020hrnet}. 
As shown in Figure~\ref{fig:hardware_motivation}, applying semantic segmentation to an input size of \( 640 \times 640 \) results in processing latencies exceeding 300 ms, with some architectures, such as Segformer, experiencing delays of over a second. This substantial processing time leads to serious latency and a noticeable drop in user experience, as achieving 10-20 frames per second (FPS), namely a processing latency of 50-100ms, is generally required for a satisfactory visual experience~\cite{albert2017latency}. In contrast, reducing the input size to \( 64 \times 64 \) greatly lowers processing latency, meeting the user requirement for speed, but on the other hand will degrade the accuracy performance.

In addition, prior studies~\cite{yang2019video, wang2021end, yan2023universal, rajivc2023segment, lin2021video, wu2022seqformer} on video instance segmentation process consecutive frames jointly (e.g., VisTR~\cite{yan2023universal} and SeqFormer~\cite{wu2022seqformer} process 5 consecutive frames together). This method leverages temporal correlations across frames to enhance performance; However, this approach introduces significant latency because processing does not begin until all frames are available, leading to a substantial delay that hinders real-time responsiveness. For example, a Meta Ray-Ban device running at 30 FPS~\cite{rayban} would experience a delay of \(5 \times \frac{1}{30} = 167 \text{ms}\), which significantly exceeds the acceptable processing latency range of 50-100 ms~\cite{albert2017latency}.

\subsection{Gaze Behavior in AR/VR Environment}
\label{sec:bg:aria_study}

\begin{figure}[t]
\centering
\includegraphics[width=1\linewidth]{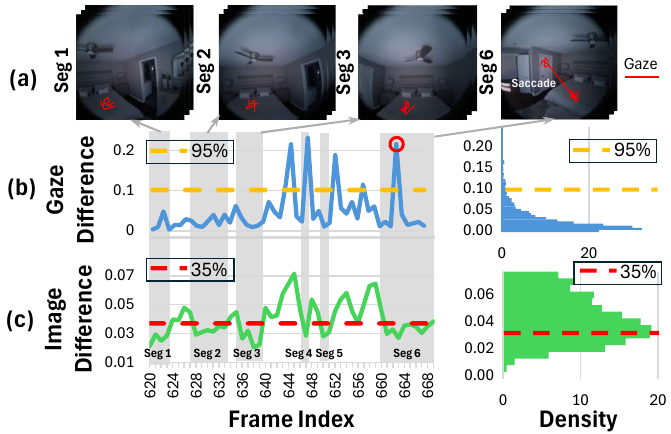}
\caption{(a) Images and corresponding gaze locations from the Aria Everyday dataset~\cite{lv2024ariaeverydayactivitiesdataset}, collected from real users wearing a VR headset. (b) Left: Changes on gaze locations over the frames. Right: Histogram of gaze differences with the yellow line marking the $95\%$ threshold. (c) Left: Normalized pixelwise differences across frames, with gray blocks indicating frames within the same segments. A \(0.037\) threshold is used to group similar frames. Right: The histogram of image differences. $35\%$ of pairs of consecutive frames with a difference less than $0.037$.}
\label{fig:aria_study}
\end{figure}
In this section, we present a detailed study of gaze behavior using real-world data from the Aria Everyday Activities Dataset~\cite{lv2024ariaeverydayactivitiesdataset}, which captures sequences of images aligned with the user's field of view and tracks gaze movement within each frame, as illustrated in Figure~\ref{fig:aria_study} (a). 

To quantify head movement, we compute the image difference by calculating the Euclidean distance between corresponding pixels of consecutive frames. If a user maintains a steady gaze in one direction for a period of time, the resulting pixel differences between frames will be minimal. When the user shifts their head, the pixel difference increases, indicating movement.
The left part of Figure \ref{fig:aria_study} (c) represents the pixel difference over a sample time interval. To identify frames with minimal changes, we set a threshold for the pixel difference. If the pixel difference falls below this threshold, the frames will be highly similar and imperceptible to the human eye, allowing them to be grouped into a single~\textbf{segment}, as shown by the gray shading in the left part of Figure \ref{fig:aria_study} (b) and (c).


Within each segment, segmentation results can be reused if the gaze remains relatively stable. To demonstrate this, we analyze gaze location difference within a segment, as shown in the left part of Figure \ref{fig:aria_study} (b). The results indicate that a threshold of \(0.1\) can be used to group gaze locations within the fixation phase; values above this threshold suggest the occurrence of a saccade, as shown in Segment 6 of Figure~\ref{fig:aria_study} (a). Additionally, \(95\%\) of frames within each segment have a gaze difference below \(0.1\). This study shows significant potential to improve the computational efficiency of instance segmentation tasks by~\textbf{focusing processing solely on the IOI and reuse the results within a segment}.


\subsection{Gaze Tracking System in AR/VR Devices}
\label{sec:bg:tfr}
Gaze tracking is pivotal in AR/VR systems, as it provides a seamless and natural interface by accurately identifying where users are focusing their attention within immersive environments. 
Most commercial AR/VR devices are equipped with integrated eye-tracking systems. For example, popular headsets such as the Meta Quest Pro~\cite{meta_quest_pro} and HTC Vive Pro Eye~\cite{hou2024unveiling} come with built-in eye trackers that monitor users' gaze in real-time, achieving latencies as low as 5-10 milliseconds and sampling rates up to 120 Hz~\cite{hou2024unveiling, stein2021comparison}. These eye-tracking modules use infrared or image sensors to capture eye movements with high accuracy and speed, enabling the system to determine where the user is looking at any given moment. 

\subsection{Image Segmentation via Input Downsampling}
\label{sec:bg:instance_segmentation}
Previous research has focused on creating efficient input downsampling methods for DNNs. By making downsampling differentiable, these methods enable the training process to adjust sampling resolution selectively, enhancing performance while reducing input dimensions and improving efficiency. In \cite{recasens2018learning}, the authors present a saliency-based distortion layer for convolutional neural networks that enhances spatial sampling of input data for image classification tasks. Subsequent works, such as \cite{jin2021learning, thavamani2021fovea, marin2019efficient}, apply similar concepts by learning a saliency score for each pixel to guide the downsampling process, resulting in improved performance for semantic segmentation tasks. However, while the zoom process is learnable, the unzoom process, which projects the label maps back to the original dimensions, is often performed using an analytical solution, leading to suboptimal results. To address this, LZU \cite{thavamani2023learning} proposes an efficient solution to learn both the zoom and unzoom processes, streamlining the entire downsampling and upsampling workflow. In contrast, FSNet utilizes user gaze input along with the captured image to perform instance segmentation only at IOI. Building on FSNet, we propose FovealSeg framework to perform instance segmentation across multiple frames with high efficiency. This approach leverages temporal correlation and human gaze behavior to optimize processing, reducing redundant computations over consecutive frames.

\section{Methodology}
\label{sec:method}
\begin{figure}[t]
\centering
\includegraphics[width=1\linewidth]{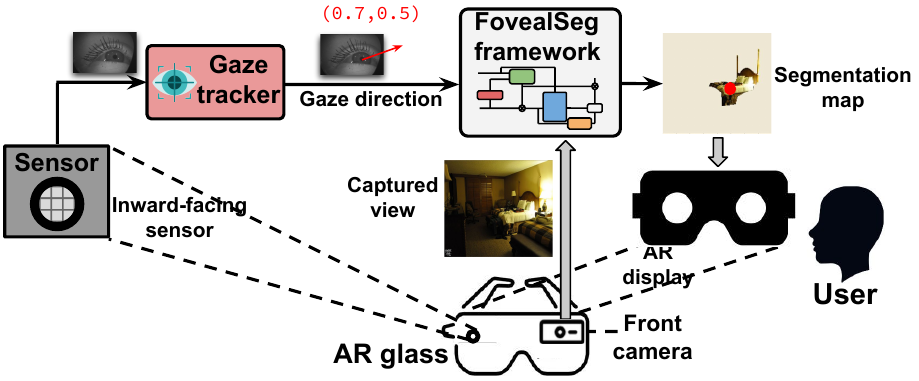}
\caption{An overview of FovealSeg framework.}
\label{fig:overview}
\end{figure}
Figure~\ref{fig:overview} illustrates the computational flow of the FovealSeg framework. During operation, the inward-facing sensor of the AR/VR device continuously captures images of the user's eye and sends them to the gaze tracker, which estimates the gaze direction with high accuracy in approximately 5-10 milliseconds~\cite{stein2021comparison, hou2024unveiling}. The estimated gaze direction is then passed to FovealSeg, along with the captured high-resolution image from the front camera as an additional input. FovealSeg generates a segmentation map focused solely on the IOI and reuse across frames with similar gaze locations. 

In this section, we start by outlining the preliminary of image sampling in Section~\ref{sec:prelim}, then provide a detailed description of the FSNet design in Section~\ref{sec:training}, and describe the FovealSeg framework in Section~\ref{sec:scheduling}.

\begin{figure*}[t]
\centering
\includegraphics[width=1\linewidth]{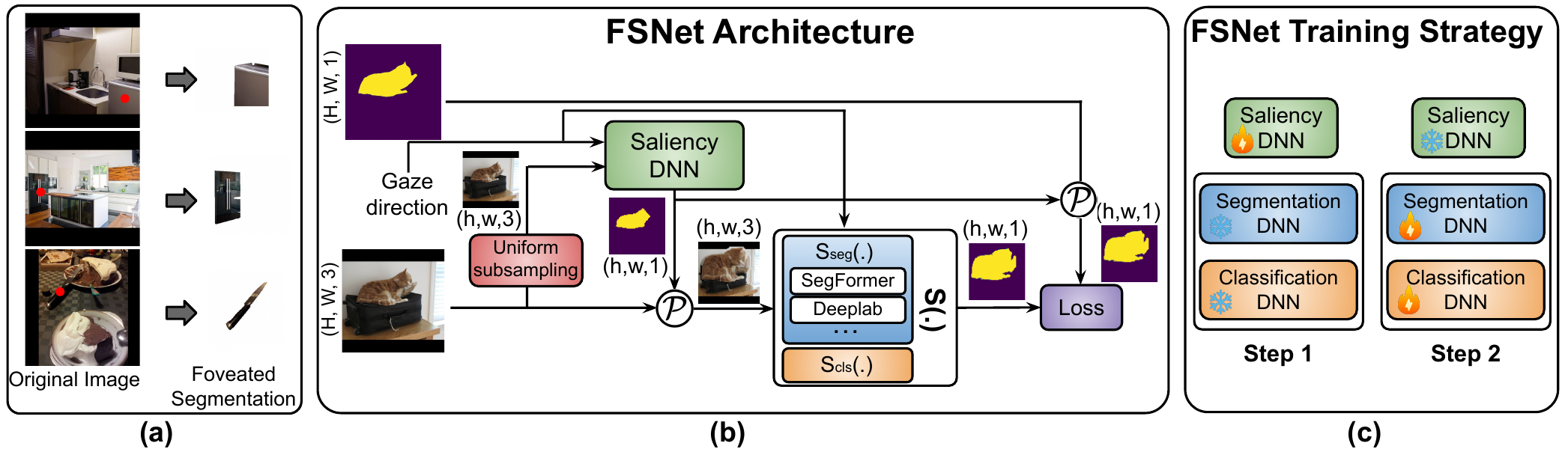}
\caption{(a) An illustration of foveated instance segmentation, where the red point indicates the gaze focus region. In this approach, only the object within the gaze area is segmented, while all other regions are considered background. (b) Overview of the FSNet architecture during the training phase. $\mathcal{P}$ denotes the saliency-guided downsampling. (c) Alterative training strategy for FSNet.}


\label{fig:fsnet_overview}
\end{figure*}

\subsection{Preliminary}
\label{sec:prelim}
Image downsampling can be viewed as an operation that transforms the original image \( F \in \mathbb{R}^{H \times W \times C} \) into a new image \( \hat{F} \in \mathbb{R}^{h \times w \times C} \), where \( h \leq H \) and \( w \leq W \). This downsampling process is achieved using two mapping functions, \( g^{h}(.) \) and \( g^{w}(.) \), which take the 2D coordinate \( (i, j) \) of the downsampled image \( \hat{F} \) and produce the corresponding coordinate \( (g^{h}(i, j), g^{w}(i, j)) \) in the input image \( F \). Thus, each pixel in \( \hat{F} \) can be defined as:
\begin{equation}
\label{eqn:mapping}
\hat{F}[i, j] := F[g^h(i, j), g^w(i, j)]
\end{equation}
where $F[i,j]$ denotes the pixel at coordinate $(i,j)$ within $F$. Moreover, to eliminate the impact of the image dimension, instead of generating the actual coordinates, the mapping functions can process the normalized 2D coordinates. For example, let $G=\{G^{h}, G^{w}\}$ denote the set of mapping function, where both $G^{h}$ and $G^{w}$ takes a normalized 2D coordinate of~$\hat{F}$, and produce normalized height and weight of X, respectively. Equation~\ref{eqn:mapping} can be rewritten as:
\begin{equation}
\label{eqn:mapping_norm}
\small
\hat{F}[i, j] := F\Biggr[ \left\lceil G^h\Bigl(\frac{i}{h}, \frac{j}{w} \Bigr) H \right\rfloor, \left\lceil G^w\Bigl(\frac{i}{h}, \frac{j}{w}\Bigr) W \right\rfloor  \Biggr]
\end{equation}
where $\lceil\cdot\rfloor$ is the rounding function, $i\leq h$ and $j\leq w$. For instance, for uniform dowmsampling function, $G^{h}, G^{w}$ are defined as follows:
\begin{equation}
\label{eqn:mapping_uniform}
\small
\hat{F}[i, j] := F\Biggr[ \left\lceil \frac{i}{h}\times H \right\rfloor, \left\lceil \frac{j}{w}\times W \right\rfloor  \Biggr]
\end{equation}

In saliency-guided downsampling~\cite{jin2021learning,thavamani2023learning}, the mapping operations G becomes learnable by incorporating a saliency-based sampling layer. In this layer, the saliency score, representing the relative sampling density at each normalized pixel location \((i, j)\), is defined by $D_{\theta}(i,j)$. Here, $D_{\theta}(.)$ is a DNN that takes the original input image \(F\) and the normalized coordinates \((i, j)\) to generate a score map with dimensions \(\mathbb{R}^{h \times w \times 1}\).
The parameter set \(\theta\) represents the weights of the DNN. Given this, the mapping function can be determined as:

\begin{equation}
\label{equ:downsample_x}
\small
G^{h}(i, j) = \frac{\sum_{i', j'} D_{\theta}(i', j') k_{\sigma}((\frac{i}{h}, \frac{j}{w}), (\frac{i'}{H}, \frac{j'}{W})) i'}{\sum_{i', j'}  D_{\theta}(i', j') k_{\sigma}((\frac{i}{h}, \frac{j}{w}), (\frac{i'}{H}, \frac{j'}{W}))}
\end{equation}

\begin{equation}
\label{equ:downsample_y}
\small
G^{w}(i, j) = \frac{\sum_{i', j'}  D_{\theta}(i', j') k_{\sigma}((\frac{i}{h}, \frac{j}{w}), (\frac{i'}{H}, \frac{j'}{W})) j'}{\sum_{i', j'} D_{\theta}(i', j') k_{\sigma}((\frac{i}{h}, \frac{j}{w}), (\frac{i'}{H}, \frac{j'}{W}))}
\end{equation}
where \(k_{\sigma}(x, x^{'})\) is a Gaussian kernel with a standard deviation of $\sigma$. To project \(\hat{F}\) back to its original size, a reverse sampler \(G^{-1}\) restores the downsampled image \(\hat{F}\) to the original space, using an interpolation function to compute the missing pixel values.

\subsection{FSNet Training Methodology}
\label{sec:training}
Foveated instance segmentation poses two main challenges: (1) As illustrated in Figure~\ref{fig:fsnet_overview} (a), foveated segmentation focuses on generating an instance segmentation mask solely for the IOI. Consequently, it requires creating a binary mask to identify the IOI region along with its associated class label.
This differs from conventional segmentation approaches, which are designed for multi-class outputs. (2) Existing methods have difficulty utilizing gaze location as prior information to guide segmentation, complicating the task of distinguishing between foreground and background across various classes. Additionally, commonly used segmentation loss functions, such as standard joint loss and focal loss~\cite{jin2021learning, lin2018focallossdenseobject}, often fall short in fine-tuning when the target instance is particularly small. To address these issues, we introduce FSNet, which incorporates (1) a gaze-aware model architecture and (2) an optimized training and fine-tuning strategy.

\paragraph{Gaze-aware Model Architecture}  
The FSNet architecture, as depicted in Figure~\ref{fig:fsnet_overview} (b), begins with an input image \( F \in \mathbb{R}^{H \times W \times 3} \). Incorporating a specified gaze location \((u, v)\), we construct a~\textit{gaze map} $N$ which is then concatenated to the input along the channelwise dimension. The map represents the normalized inverse distance to the gaze point, with higher values (closer to 1) signifying closeness to the gaze location, each element $N[i,j]$ of $N$ is defined as follows:
\[
N[i,j] = 1 - \frac{\sqrt{(i - u)^2 + (j - v)^2}}{d_{\text{max}}}
\]
Here, \(0 \leq i \leq H\) and \(0 \leq j \leq W\). The term \(d_{\text{max}} = \sqrt{H^2 + W^2}\) represents the maximum possible distance between any two pixels in \(F\), serving as a normalization factor. We concatenate the gaze map with the input image and downsample the result as input to the saliency DNN. The saliency DNN produces a saliency map \(D_{\theta}(i,j) \in \mathbb{R}^{H \times W \times 1}\) for each coordinate \(1 \leq i \leq H\) and \(1 \leq j \leq W\). This score map will then guide the sampling of the input image \(F\) using equations~\ref{equ:downsample_x} and~\ref{equ:downsample_y}, producing \(\hat{F}\). The resulting \(\hat{F}\in \mathbb{R}^{h \times w \times 3}\) will have an enlarged IOI region to enhance the performance of the instance segmentation operation conducted by the segmentation network \(S(.)\), which produce segmentation mask $Y\in \mathbb{R}^{h \times w \times 1}$.


The problem of foveated instance segmentation differs from conventional segmentation in that we only need the segmentation map for the IOI, not the entire image. Therefore, it is unnecessary for the segmentation network \(S(.)\) to produce a pixel-wise output for every object within $F$. Instead, we modify the segmentation network to include two branches, \(S = \{S_{\text{seg}}(.), S_{\text{cls}}(.)\}\), as illustrated in Figure~\ref{fig:fsnet_overview} (b). 
The first branch, \(S_{\text{seg}}(.)\), produces a binary map \(Y_{\text{bm}} \in \mathbb{R}^{h \times w \times 1}\) to represent the IOI mask, where elements of \(Y_{\text{bm}}\) is set to 1 for regions belonging to the IOI and 0 otherwise. The second branch, \(S_{\text{cls}}(.)\), classifies the object within the IOI, yielding an output \(Y_{\text{cls}} \in \mathbb{R}^{C \times 1}\), where $C$ is the number of possible classes. The final segmentation label \(Y \in \mathbb{R}^{h \times w \times 1}\) is then produced by performing an outer product between \(Y_{\text{cls}}\) and \(Y_{\text{bm}}\), producing the final segmentation mask of \(Y_{\text{cm}} \in \mathbb{R}^{h \times w \times C}\)
This design reduces the amount of output generated by the segmentation network and simplifies the training process by leveraging the characteristics of the foveated segmentation task.

In practice, \( S_{\text{seg}} \) can be any pretrained neural network designed for segmentation tasks, such as SegFormer~\cite{xie2021segformer}, DeepLab~\cite{chen2017deeplab}, among others. During the inference phase, the $Y_{\text{cm}}$ is upsampled using the deterministic interpolication process based on the sampler $G^{h}$ and $G^{w}$.

\paragraph{Loss Design} To compute the loss, we follow a methodology of~\cite{jin2021learning}, where the ground truth mask $Y_{gt}\in \mathbb{R}^{H \times W \times C}$ is subsampled using the identical saliency score map $D_{\theta}(i,j)$ as the input F, resulting in a subsampled version of the ground truth map $Y^{'}_{gt}\in \mathbb{R}^{h \times w \times C}$. The Dice loss $\mathbb{L}_{dice}$ is then computed between $Y^{'}_{gt}$ and $Y_{\text{cm}}$. 
Additionally, a unique characteristic of foveal instance segmentation is that the IOI is sometimes quite small (e.g., objects like a knife or pot, as shown in Figure~\ref{fig:fsnet_overview} (a)). Consequently, when calculating the pixelwise loss function, if pixels within the IOI are weighted equally with those in the non-IOI region, the results tend to be heavily biased toward the segmentation mask of the non-IOI region, leading to an unintended emphasis on non-IOI regions. 
To mitigate this, we weigh the pixelwise Focal loss by the inverse of the area of the IOI and the non-IOI region, specifically, the loss $\mathbb{L}_{tot}$ can be defined as:
\begin{equation}
    \label{eqn:loss}
    \mathbb{L}_{tot} = L_{dice}(Y^{'}_{gt},Y_{\text{cm}}) + \lambda L_{focal}(Y^{'}_{gt},Y_{\text{cm}})
\end{equation}
where, $\mathbb{L}_{dice}(.)$ and $\mathbb{L}_{focal}(.)$ are the dice loss and weighted focal loss functions, respectively. $\lambda$ denotes the relative importance between $\mathbb{L}_{dice}(.)$ and $\mathbb{L}_{focal}(.)$.

\paragraph{Alternative Training Strategy} 
To train FSNet, we employ an alternate training strategy, as illustrated in Figure~\ref{fig:fsnet_overview} (c). Initially, the segmentation neural network components, including $S_{\text{seg}}(.)$ and $S_{\text{cls}}(.)$, are kept frozen while the saliency DNN undergoes training for several epochs. Subsequently, the saliency DNN is frozen, and the segmentation neural network is fine-tuned with a distinct learning rate.

\subsection{FovealSeg Framework}
\label{sec:scheduling}

\begin{algorithm}[t]
\caption{FovealSeg Algorithm}\label{alg:fovealseg}
\small
\KwIn{$T$ is the total time. $F^{init}$ is the initial frame of current video segment. $g_{last}$ and $M_{last}$ are buffered gaze location and segmentation mask. $g_{t}$, $F^{t}$ and $M_{t}$ are current gaze location, input frame and segmentation mask. Threshold $\alpha$ and $\beta$ for the detection of saccade and end of segment.}
\SetKwBlock{Begin}{Initiation}{}
\Begin{
    $F^{init} = \varnothing$, $g_{last} = \varnothing$, $M_{last} = \varnothing$\\
    \For{$1\leq t \leq T$}{
    \If{$|g_{t}-g_{last}|^{2} > \alpha$}{
        $g_{last} \gets g_{t}$\;
        Saccade detect, halt rest operations.
    }
    \Else{
        \If{$\sum_{ij}|F^{t}_{ij}-F^{init}_{ij}| > \beta$}{
            Run FSNet with $F^{t}$ and $g_{t}$, get $M^{t}$\;
            $F^{init} \gets F^{t}$, $g_{last} \gets g_{t}$, $M_{last} \gets M_{t}$\;
            \KwRet $M_{t}$
        }
        \Else{
            \If{$g_{t}$ \text{is within IOI regions of} $M_{last}$}{
                \KwRet $M_{last}$
            }
            \Else{
                Run FSNet with $F^{t}$ and $g_{t}$, get $M^{t}$\;
                $g_{last} \gets g_{t}$, $M_{last} \gets M_{t}$\;
                \KwRet $M_{t}$
            }
        }
    }
    }
}
\end{algorithm}
Building on the description of FSNet in Section~\ref{sec:training}, this section discusses how it integrates into the FovealSeg framework to enable efficient detection across multiple frames. The FovealSeg algorithm is outlined in Algorithm~\ref{alg:fovealseg}. Initially, a simple criterion is applied to detect the occurrence of a saccade by calculating the difference between the current and previous gaze positions (line 4). If a saccade is detected, instance segmentation for the current frame can be skipped due to the reduced sensitivity of the human visual system during a saccade (line 6). If no saccade occurs, the similarity between the current frame $F^{t}$ and the initial frame of the current segment $F^{init}$ is assessed by computing their difference. If this difference exceeds a predefined threshold $\beta$ (line 8), it indicates a significant change in the scene, triggering a full re-execution of instance detection (line 9), and record the updated segmentation mask at the gaze location (line 10). If not, the current gaze location is analyzed to determine if it remains within the region defined by the segmentation mask $M_{last}$. If it does, $M_{last}$ can be reused; otherwise FSNet must be executed with new gaze location (line 16).

\section{Evaluation}
\label{sec:experiment}
To validate the effectiveness of the FSNet and FovealSeg frameworks, we perform extensive evaluations using multiple baseline algorithms and datasets. In Section~\ref{sec:fsnet-eval}, we compare FSNet's performance against different baseline methods across three diverse datasets, emphasizing its advancements. In Section~\ref{sec:fovealseg-eval}, we provide quantitative performance results of the FovealSeg framework in AR/VR scenarios, showcasing its suitability for immersive environments. Lastly, we present ablation studies in Section~\ref{sec:ablations} to further demonstrate our method's optimality and adaptability across various conditions.

\begin{table}[t]
\centering
\resizebox{0.47\textwidth}{!}{%
\begin{tabular}{|c|c|c|c|}
\hline
\textbf{Dataset}  & \textbf{Image size} & \textbf{Number of classes} & \textbf{Size of dataset} \\ \hline
Cityscape & $512 \times 1024$ & 19 & 20000 \\ \hline
ADE20K & $640 \times 640$ & 31 & 30000 \\ \hline
LVIS & $640 \times 640$ & 50 & 100000 \\ \hline
Aria Everyday Activities & $1402 \times 1402$ & 100 & 60000 \\ \hline
\end{tabular}%
}
\caption{The overview of evaluation dataset used in our work. In each dataset, we uniformly sample instances to ensure a balanced distribution across categories.}
\label{tab:dataset}
\end{table}

\subsection{Experiment settings}
\label{sec:settings}
\begin{table*}[htbp]
\centering
\renewcommand{\arraystretch}{1}
\setlength{\tabcolsep}{6pt}
\begin{adjustbox}{width=0.77\textwidth,center}
\begin{tabular}{c|cc|cc|cc|cc}
\hline
\textbf{Method} 
& \multicolumn{2}{c|}{\small\textbf{CityScapes} $(64 \times 128)$} 
& \multicolumn{2}{c|}{\small\textbf{ADE20K} $(80 \times 80)$} 
& \multicolumn{2}{c|}{\small\textbf{LVIS} $(80 \times 80)$} 
& \multicolumn{2}{c}{\small\textbf{Aria} $(180 \times 180)$} 
\\ \cline{2-9} 
& \textbf{IoU} $\uparrow$ & \textbf{IoU}$'$ $\uparrow$ 
& \textbf{IoU} $\uparrow$ & \textbf{IoU}$'$ $\uparrow$ 
& \textbf{IoU} $\uparrow$ & \textbf{IoU}$'$ $\uparrow$  
& \textbf{IoU} $\uparrow$ & \textbf{IoU}$'$ $\uparrow$  
\\ \hline
Avg+DeepLab         & 0.26 & 0.27 & 0.39 & 0.41 & 0.35 & 0.36 & 0.36 & 0.37\\ 
Avg+HRNet           & 0.20 & 0.21 & 0.43 & 0.44 & 0.37 & 0.38 & 0.39 & 0.41 \\ 
Avg+SegFormer-B4    & 0.25 & 0.27 & 0.37 & 0.39 & 0.37 & 0.38 & 0.36 & 0.36 \\ 
Avg+SegFormer-B5    & 0.27 & 0.29 & 0.41 & 0.42 & 0.35 & 0.37 & 0.37 & 0.38\\ 
LTD~\cite{jin2021learning} & 0.37 & 0.38 & 0.41 & 0.41 & 0.40 & 0.41 & 0.41 & 0.43\\ 
\rowcolor{blue!10}
FSNet+DeepLab       & \textbf{0.52} & \textbf{0.53} & 0.55 & 0.56 & 0.53 & 0.55 & 0.54 & 0.56\\ 
\rowcolor{blue!10}
FSNet+HRNet         & 0.47 & 0.49 & \textbf{0.56} & 0.56 & \textbf{0.54} & \textbf{0.55} & 0.57 & 0.58 \\ 
\rowcolor{blue!10}
FSNet+SegFormer-B4  & 0.46 & 0.48 & 0.54 & 0.55 & 0.54 & 0.56 & 0.56 & 0.57\\ 
\rowcolor{blue!10}
FSNet+SegFormer-B5  & 0.51 & 0.52 & 0.55 & \textbf{0.57} & 0.54 & 0.55 & \textbf{0.58} & \textbf{0.59}\\ \hline
\end{tabular}
\end{adjustbox}

\caption{
Performance comparison on CityScapes, ADE20K, LVIS and Aria Everyday Activities datasets. 
}
\label{tab:performance_comparison}
\end{table*}

\paragraph{Datasets:} We utilize four publicly available datasets: Cityscapes~\cite{Cordts2016Cityscapes}, ADE20K~\cite{zhou2019ade20k}, LVIS~\cite{gupta2019lvisdatasetlargevocabulary}, and Aria Everyday Activities~\cite{lv2024ariaeverydayactivitiesdataset}. Given that these datasets are designed for full-image segmentation, we applied a gaze-aware masking preprocessing technique to enable foveated instance rendering. Due to the space limit, some of the evaluation results and the details of the preprocessing steps are outlined in the Appendix. For each training and test sample, we randomly select a gaze location within the image and define the IOI based on this gaze location. The details can be found in Table~\ref{tab:dataset}.

\paragraph{Model selection:} We use a lightweight 3-layer U-Net~\cite{ronneberger2015unet} as the saliency DNN and a pre-trained MobileNetV2~\cite{sandler2019mobilenetv2} as the classification DNN $S_{cls}(.)$. For $S_{seg}(.)$, we adopt several widely-used architectures, including DeepLabV3~\cite{chen2017deeplab}, SegFormer\footnote{Two versions are used: SegFormer-B4 and SegFormer-B5}~\cite{xie2021segformer} and HRNet~\cite{wang2020hrnet}. We modify the architectures of $S_{seg}(.)$ and $S_{cls}(.)$ to ensure alignment with the dimensions of the downsampled input image $F'$ and the distance map $C$.
Two baselines are developed. In the first baseline, the algorithm uniformly subsamples the input $F$ to produce $F'$ instead of using the saliency DNN. The resulting $F'$ is then processed by the segmentation neural network $S(.)$, with $S_{seg}(.)$ and $S_{cls}(.)$ remaining identical to those in FSNet, following the same training strategy. We also compare FSNet with Learn-to-Downsample (LTD)~\cite{jin2021learning}, which uses a learnable downsampling approach combined with an edge-based loss function to perform instance segmentation on the entire frame. 



\paragraph{Evaluation metrics:} We use $\text{IoU}$ and IoU\textquotesingle~to evaluate instance segmentation performance over IOI within both the full-resolution test input $F$ and its subsampled version $F'$. Although LTD performs segmentation across the entire frame, we focus only on the IoU within the IOI where the gaze is directed.

\begin{table*}[h]
    \centering
    \centering\resizebox{1.9\columnwidth}{!}{
    \begin{minipage}{0.32\textwidth}
    \renewcommand{\arraystretch}{0.85}
    \setlength{\tabcolsep}{4pt}
        \centering
        \begin{tabular}{c|c|c}
            \toprule
            \textbf{Method} & \textbf{IoU$\uparrow$} & \textbf{IoU\textquotesingle$\uparrow$} \\
            \midrule
            Avg+DeepLab   & 0.16  & 0.17 \\
            Avg+HRNet  & 0.16  & 0.17 \\
            Avg+Seg-B4  & 0.16  & 0.18 \\
            Avg+Seg-B5 & 0.16  & 0.18 \\
                        \rowcolor{green!10}
            FSNet+DeepLab   & 0.36  & 0.37 \\
            \rowcolor{green!10}
            FSNet+HRNet   & 0.29  & 0.36 \\
                        \rowcolor{green!10}
            FSNet+Seg-B4   & 0.32  & 0.35 \\
                        \rowcolor{green!10}
            FSNet+Seg-B5 & 0.36  & 0.38 \\
            \bottomrule
        \end{tabular}
        \caption{Influence of downsample rate on CityScapes (low resolution $32 \times 64$).}
        \label{tab:downsample_rate}
    \end{minipage}%
    \hfill
    \begin{minipage}{0.35\textwidth}
    \renewcommand{\arraystretch}{0.85}
    \setlength{\tabcolsep}{4pt}
        \centering
        \begin{tabular}{c|c|c|c}
            \toprule
            \textbf{Method} & \textbf{Kernel Size} & \textbf{IoU $\uparrow$}&\textbf{IoU\textquotesingle$\uparrow$} \\
            \midrule
            DeepLab   &17 &0.48  & 0.49 \\
            HRNet  &17 &0.45  & 0.46 \\
            Seg-B4&17  & 0.41  & 0.44 \\
            Seg-B5 &17 &0.47  & 0.47 \\
                        \rowcolor{green!10}
            DeepLab   &33 &0.52  & 0.53 \\
                        \rowcolor{green!10}
            HRNet   & 33&0.47  & 0.49 \\
                        \rowcolor{green!10}
            Seg-B4& 33  & 0.46  & 0.48 \\
                        \rowcolor{green!10}
            Seg-B5& 33& 0.51  & 0.52 \\
            \bottomrule
        \end{tabular}
        \caption{Effect of gaussian kernel size deployed by sampler on FSNet.}
        \label{tab:kernel_size}
    \end{minipage}%
    \hfill
    \begin{minipage}{0.32\textwidth}
    \renewcommand{\arraystretch}{0.85}
    \setlength{\tabcolsep}{4pt}
        \centering
        \begin{tabular}{c|c|c}
            \toprule
            \textbf{Method}  & \textbf{IoU $\uparrow$}&\textbf{IoU\textquotesingle$\uparrow$} \\
            \midrule
            DeepLab(w/o)   &0.14  & 0.15 \\
            HRNet(w/o)   &0.15  & 0.17 \\
            Seg-B4(w/o)  & 0.15  & 0.16 \\
            Seg-B5(w/o)  &0.16  & 0.16 \\
                        \rowcolor{green!10}
            DeepLab(w/)    &0.52  & 0.53 \\
                        \rowcolor{green!10}
            HRNet(w/)   &0.47  & 0.49 \\
                        \rowcolor{green!10}
            Seg-B4(w/)  & 0.46  & 0.48 \\
                        \rowcolor{green!10}
            Seg-B5(w/)& 0.51  & 0.52 \\
            \bottomrule
        \end{tabular}
        \caption{Influence of gaze information on FSNet performance.}
        \label{tab:gazeinformation}
    \end{minipage}
    }
\end{table*}

\subsection{Evaluation Results of FSNet}
\label{sec:fsnet-eval}
Table~\ref{tab:performance_comparison} presents a summary of our results. Among the methods, Avg+DeepLab refers to the uniformly subsampling baseline with a pretrained DeepLab as $S_{seg}(.)$, and similarly for other architectures. Likewise, FSNet+DeepLab represents FSNet with a pretrained DeepLab as $S_{seg}(.)$. For ADE20K and LVIS, the input images are uniformly downsampled to $(80, 80)$, while for Cityscapes and Aria Everyday Activities datasets, the input images are downsampled to $(64, 128)$ and $(180, 180)$, respectively.

As shown in the results, FSNet consistently outperforms the baseline across all datasets and four segmentation backbones, achieving at least a 0.14 improvement in IoU and a 0.15 gain in IoU\textquotesingle~compared to the Avg method. For instance, FSNet combined with SegFormer-B5 achieves an IoU of 0.58 and an IoU\textquotesingle~of 0.59, surpassing all baselines. Furthermore, FSNet demonstrates strong generalizability across datasets, although slight performance variations are observed due to dataset-specific characteristics. The small gap between IoU and IoU\textquotesingle~($<0.02$) suggests that our stage-aware training strategy effectively improves the robustness of both upsampling and downsampling stages.

\subsection{Evaluation Results on FovealSeg Framework}
\label{sec:fovealseg-eval}


\begin{figure}[t]
\centering
\includegraphics[width=0.85\linewidth]{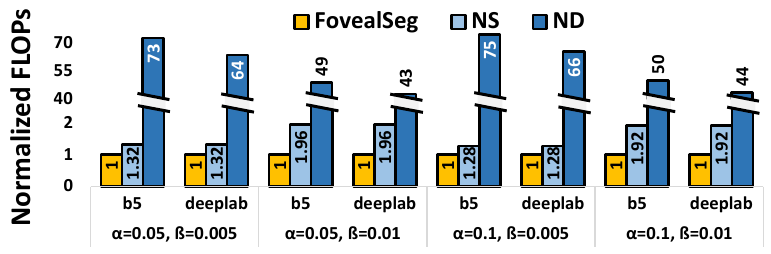}
\caption{Normalized number of FLOPs of the FovealSeg framework compared to the baselines across different models and combinations of $\alpha$ and $\beta$.}
\label{fig:ff_eval}
\end{figure}
The trained FSNet can then be integrated into the FovealSeg framework, as outlined in Algorithm~\ref{alg:fovealseg}, to process segments. To assess the system's performance improvement, we use consecutive frames from the Cityscapes dataset. However, since the Cityscapes dataset lacks gaze location data, we incorporate gaze traces from the Everyday Activity dataset into the Cityscapes dataset. To evaluate FovealSeg, we develop two baseline algorithms. The first baseline algorithm, called~\textbf{No downsample (ND)}, replaces FSNet with a conventional segmentation DNN that processes the full-resolution input image at a resolution of \(640 \times 640\). This baseline aims to evaluate the impact of the downsampling operation on computational efficiency. The second baseline applies FovealSeg framework processes images at the downsampled resolution of $64 \times 64$ without frame skipping, referred to as \textbf{No Skip (NS)}, aim to show the importance of gaze reuse across frames. We configure FovealSeg under different settings for $\alpha$ and $\beta$ for saccade detection and segment detection, respectively. All the $\alpha$ and $\beta$ are set to ensure similar frames are grouped and have negligible impact to the model accuracy.


The results in Figure~\ref{fig:ff_eval}, highlight the enhanced computational efficiency achieved by FovealSeg framework compared to the baselines. The high downsampling rate employed by FovealSeg framework reduces the computation required for instance segmentation tasks. As discussed in Section~\ref{sec:bg:aria_study}, leveraging gaze saccades and fixations allows for the elimination of a significant amount of redundant computations, achieving up to a $1.96\times$ reduction in FLOPs. Compared with ND, FovealSeg can achieve up to $75\times$ reduction in computation, underscoring substantial contribution of downsampling to system performance enhancement.

\subsection{Ablation Studies}
\label{sec:ablations}
\textbf{Influence of the Downsample Rate.} We start by examining the impact of the downsampling rate in FSNet on performance. Table~\ref{tab:downsample_rate} presents FSNet performance with a downsampled image size $F'$ of $32 \times 64$ on the CityScapes dataset. Both IoU and IoU\textquotesingle~ show a notable decline as the downsampling ratio increases, with the IoU for the DeepLab-based FSNet dropping from 0.52 to 0.36. This trend is consistent across all baseline methods. Nevertheless, even at this low resolution, our FSNet method still outperforms the others.\\
\textbf{Influence of the Gaussian Kernel Size in Sampler.} We examine how the Gaussian kernel size $\sigma$ used in equations~\ref{equ:downsample_x} and~\ref{equ:downsample_y} affects performance, as it plays a crucial role in the sampling process. Table~\ref{tab:kernel_size} presents the results on the CityScapes dataset. The results indicate that a larger kernel size yields improved outcomes due to the increased emphasis on the saliency region.\\
\textbf{Influence of the Gaze Information.} In FSNet design, we incorporate gaze coordinates $(u, v)$ to guide FSNet in sampling the input image $F$. In Table~\ref{tab:gazeinformation}, we evaluate FSNet's performance on the CityScapes dataset without gaze information by substituting it with random noise.  The results show a clear decrease in IoU by over 0.3, highlighting the critical role of gaze location information in FSNet.

\subsection{System Performance and Visual Experience}
\label{sec:latency_miou}

To assess the real-world efficiency of our FovealSeg framework, we compare FSNet+Seg-B5 and SegFormer-B5 to evaluate the speed-accuracy trade-off. To simulate the system performance, we use GPGPU-Sim~\cite{Khairy2018AccelSimAE} to run the image segmentation models. GPGPU-Sim~\cite{Khairy2018AccelSimAE} is configured to simulate the Jetson Orin NX~\cite{jetsonorin}, a widely used edge GPU adopted by ARVR devices~\cite{zhang2024boxr, zhang2024co, pancrisp}. As illustrated in Figure~\ref{tab:latency_only_comparison}, FSNet achieves a latency of 84\,ms, making it over \textbf{20$\times$} faster than SegFormer-B5, which requires 1860\,ms. Notably, FSNet also delivers superior segmentation performance. Its lower latency leads to improved temporal alignment and better visual experience.

\begin{table}[t]
\centering
\renewcommand{\arraystretch}{1.1}

\begin{tabular}{|>{\centering\arraybackslash}m{1.8cm}|
                >{\centering\arraybackslash}m{1.05cm}|
                >{\centering\arraybackslash}m{0.06\textwidth}|
                >{\centering\arraybackslash}m{0.06\textwidth}|
                >{\centering\arraybackslash}m{0.06\textwidth}|}
\hline
\textbf{Model} & \textbf{Latency} & \textbf{150} & \textbf{151} & \textbf{152} \\
\hline
SegFormer-B5 & 1860\,ms &
\adjustimage{width=\linewidth,valign=m}{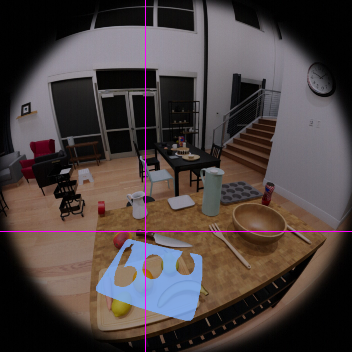} &
\adjustimage{width=\linewidth,valign=m}{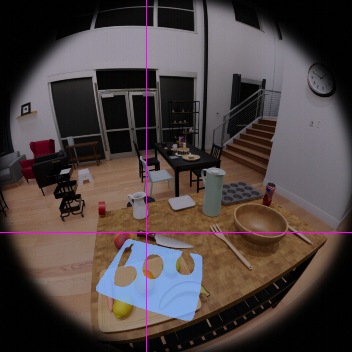} &
\adjustimage{width=\linewidth,valign=m}{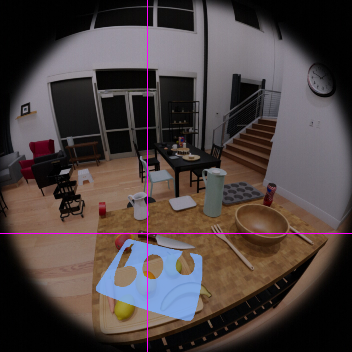} \\
\hline
FovealSeg & 84\,ms &
\adjustimage{width=\linewidth,valign=m}{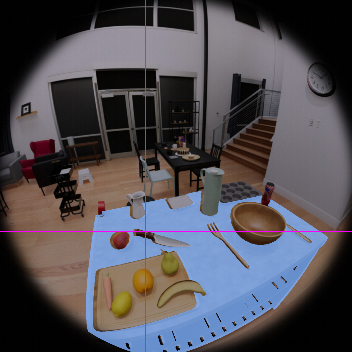} &
\adjustimage{width=\linewidth,valign=m}{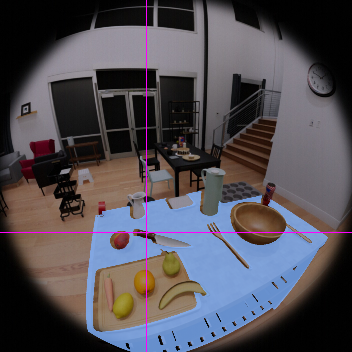} &
\adjustimage{width=\linewidth,valign=m}{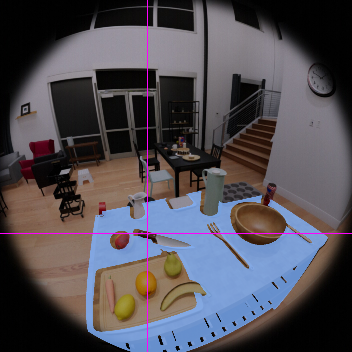} \\
\hline
\end{tabular}

\caption{Latency and qualitative analysis of sequentially chosen frames (150-152). Due to significant processing latency, SegFormer-B5 may result in a delay between the current gaze location and the predicted segmentation mask.}
\label{tab:latency_only_comparison}
\end{table}

\section{Conclusion}
 We present the FovealSeg, which leverages real-time gaze data for instance segmentation focused on the region of IOI. The evaluation results show enhanced performance across various datasets, coupled with notable efficiency improvements, setting the stage for promising future research.
{
    \small
    \bibliographystyle{ieeenat_fullname}
    \bibliography{main}
}


\end{document}


\clearpage
\setcounter{page}{1}
\maketitlesupplementary

\section{Dataset}
In the evaluation section, we evaluate our method on three segmentation datasets: CityScapes~\cite{Cordts2016Cityscapes}, ADE20K~\cite{zhou2019ade20k} and LVIS~\cite{gupta2019lvisdatasetlargevocabulary}

The \textbf{Cityscapes dataset} consists of 5000 high-resolution (2048 $\times$ 1024 pixels) images depicting urban scenes from 27 different cities across Europe. This dataset includes finely-annotated images covering 30 classes, out of which 19 are designated for training and evaluation, as per the guidelines. The 5000 images are split into 2975 for training, 500 for validation, and 1525 for testing.

The \textbf{ADE20K} dataset is a large-scale, high-quality dataset for scene parsing, containing over 20000 images that span a wide range of scenes, objects, and activities. Each image is densely annotated with pixel-level labels, covering 150 distinct object categories, including buildings, animals, and vehicles. The dataset is split into 20210 images for training, 2000 for validation, and 3352 for testing. Due to its extensive variety and meticulous annotations, ADE20K is widely used for benchmarking in semantic segmentation and scene understanding tasks.

The \textbf{LVIS} dataset is a comprehensive dataset designed for instance segmentation, containing over 1000000 high-quality instance segmentations across a wide array of 1203 object categories. Each image in LVIS is annotated with precise masks and instance-level labels, making it suitable for tasks that require detailed object segmentation and recognition. The dataset is divided into training, validation, and testing sets, with the training set containing around 100000 images. LVIS is often used for advancing research in instance segmentation, handling rare classes, and addressing challenges posed by the diversity and imbalance of real-world data.
\section{Experiment Setting}

\subsection{Dataset Pre-Processing}
\label{sec:preprocessing}
\begin{figure}[t]
\centering
\includegraphics[width=1\linewidth]{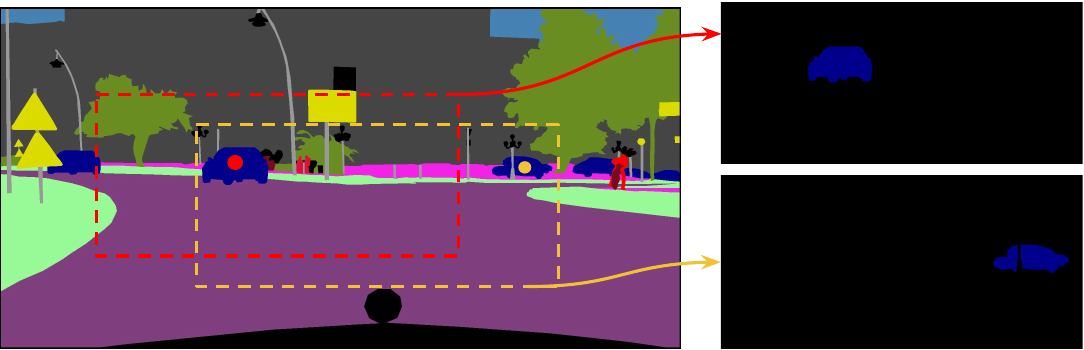}
\caption{An illustration of the Pre-Processing in our work. The solid circle denotes the gaze location, we map the object gaze at as the ground truth, while all other objects as the background.}
\label{fig:overview}
\end{figure}

To validate the performance of our FovealSeg framework, we applied a gaze-aware pre-processing technique to the original dataset. Specifically, we use gaze information as a focal point in each image. The object being gazed at is marked as the ground truth, with the region of this object set to '1' in the ground truth mask using a polygon filter. All other objects are treated as background and assigned a value of '0'. This pre-processing helps distinguish foreground from background in the image, similar to the approach used in foveated instance segmentation. We applied this same pre-processing method to the CityScapes, ADE20K, and LVIS datasets. 

This pre-processing approach enables us to create an effectively unlimited training and validation dataset by sampling from countless possible gaze locations. We utilize gaze movement trajectories from the OpenEDS2020 dataset~\cite{palmero2020openeds2020openeyesdataset} to sample gaze locations, applying them uniformly across the original dataset while maintaining a balanced distribution across object classes. This ensures equal representation for each object class, strengthening the robustness of the data. Specifically, we evaluate all 19 classes in CityScapes by cropping the original images to $512 \times 1024$. For ADE20K and LVIS, containing 31 and 50 classes respectively, we resize the images to $640 \times 640$ through padding.

\begin{table}[h]
\renewcommand{\arraystretch}{1}
\centering
\resizebox{0.47\textwidth}{!}{%
\begin{tabular}{|c|c|c|c|}
\hline
\textbf{Model}  & \textbf{Parameters} & \textbf{Version} & \textbf{Weight Source} \\ \hline
DeepLab & 42M & ResNet50 & torchvision~\cite{torchvision2016} \\ \hline
PSPNet & 24.3M & ResNet50 & torchvision \\ \hline
HRNet & 67.12M  &W48 & huggingface~\cite{wolf2020huggingfacestransformersstateoftheartnatural} \\ \hline
SegFormer-B4 & 64M & B4 & huggingface \\ \hline
SegFormer-B5 & 84.6M & B5 & huggingface \\ \hline
\end{tabular}%
}
\caption{The overview of evaluation dataset used in our work. In each dataset, we uniformly sample instances to ensure a balanced distribution across categories.}
\label{tab:pretrained}
\end{table}

\subsection{Pretrained Model Selection}
We structure the FSNet segmentation module by utilizing the pretrained models. The details can be found in table~\ref{tab:pretrained}. For the deformation module, we apply a simple U-Net structure with scalability. For the deformation module, we utilize a scalable U-Net architecture that can be adjusted based on the task requirements. Here we set the base channel to 16 and the model depth to 3 with a classifier head to generate the mask map.

\subsection{Training Details}
We used the 3-layers U-Net with a base channel of 16 as the saliency DNN and train the entire model on a server with 4 RTX6000 GPU. In the training process, we follow the alternative training strategy we proposed in the paper. During training, we applied the pre-processingand apply  data normalization on each seperate channel.

In the first training stage, the segmentation DNN and classification DNN were frozen, and only the saliency DNN was trained to minimize the loss \(\mathbb{L}(\hat{Y}, Y)\). During this stage, the weights of the saliency DNN were randomly initialized, and the models were trained using the NAdam optimizer for 500 iterations with a weight decay of \(1 \times 10^{-5}\) to prevent overfitting. An early stopping mechanism with a patience of 20 epochs was also employed. Specifically, a batch size of 128 was used for the CityScapes dataset, while batch sizes of 256 were applied to the ADE20K and LVIS datasets. The learning rate was initialized to 0.05, and a \textit{ReduceLROnPlateau} learning rate schedule with a decay factor of 0.9 was adopted.

In the second training stage, we fine-tuned the segmentation DNN and classification DNN while keeping the saliency DNN frozen. The model was initialized with pretrained weights, as detailed in Table~\ref{tab:pretrained}. Training was conducted using the AdamW optimizer over 800 iterations, with a weight decay of \(1 \times 10^{-5}\) to mitigate overfitting. For the CityScapes dataset, a batch size of 80 was used, whereas batch sizes of 120 were employed for the ADE20K and LVIS datasets. The learning rates were set to \(5 \times 10^{-3}\) for DeepLab, PSPNet, and HRNet, and \(5 \times 10^{-1}\) for Segformer. A \textit{ReduceLROnPlateau} learning rate schedule with a decay factor of 0.9 was applied. The foveated instance segmentation performance was evaluated using Intersection over Union (IoU) metrics on the focus object.

\subsection{Evaluation Details}
To evaluate the performance of FSNet on different datasets, we generate validation sets with a size equal to \(\frac{1}{5}\) of the training dataset, following the pre-processing procedure described in Section~\ref{sec:preprocessing}. Specifically, for CityScapes, the validation dataset consists of 4000 images across 19 classes; for ADE20K, it contains 6000 images across 31 classes; and for LVIS, it includes 20,000 images across 50 classes.

\section{Effect of Gaussian Kernel}
Following prior work~\cite{jin2021learning, recasens2018learning}, we adopt a Gaussian kernel-based mapping function in the process of saliency-guided downsampling. The mapping function can be expressed as:
\begin{equation}
\label{equ:downsample_x}
\small
G^{h}(i, j, F) = \frac{\sum_{i', j'} D_{\theta}(i', j', F) k_{\sigma}((i, j), (i', j')) i'}{\sum_{i', j'} D_{\theta}(i', j', F) k_{\sigma}((i, j), (i', j'))}
\end{equation}

\begin{equation}
\label{equ:downsample_y}
\small
G^{w}(i, j, F) = \frac{\sum_{i', j'} D_{\theta}(i', j', F) k_{\sigma}((i, j), (i', j')) j'}{\sum_{i', j'} D_{\theta}(i', j', F) k_{\sigma}((i, j), (i', j'))}
\end{equation}
Here, \(k_{\sigma}(x, x^{'})\) represents a Gaussian kernel with a standard deviation \(\sigma\).
\subsection{Gaussian Kernel Principle}

The Gaussian kernel is a symmetric, localized function used to weight neighboring data points based on their distances from the center. It is defined mathematically as:

\begin{equation}
\label{equ:gaussian_kernel}
k_{\sigma}((i, j), (i', j')) = \exp\left(-\frac{\| (i, j) - (i', j') \|^2}{2\sigma^2}\right),
\end{equation}

where \((i, j)\) represents the coordinates of the center pixel, \((i', j')\) are the surrounding pixel coordinates, \(\| (i, j) - (i', j') \|\) is the Euclidean distance between them, and \(\sigma\) controls the spread of the kernel. A smaller \(\sigma\) results in a sharper kernel with more emphasis on nearby pixels, while a larger \(\sigma\) produces a smoother kernel, distributing weights more broadly.

\subsection{Gaussian Kernel in FSNet}
\label{sec:kernel size}
The Gaussian kernel used in FSNet is defined with a fixed standard deviation \(\sigma\) and a square shape of size \(2\sigma+1\). To ensure proper alignment during convolution, the kernel is applied with padding of \(\sigma\), preserving the spatial dimensions of the feature map. This configuration enables the kernel to emphasize the local region around each pixel while maintaining the consistency of input and output dimensions.

According to Equation~\ref{equ:gaussian_kernel}, a small \(\sigma\) causes the weights to concentrate on the central region, effectively diminishing the contributions of pixels farther from the center. In contrast, a larger \(\sigma\) reduces the weight differences between pixels, creating a more uniform distribution of Gaussian weights. The kernel size, however, also plays a crucial role. A small kernel size limits the receptive field, focusing only on a local region and potentially losing the capacity to magnify the area of interest. Conversely, a larger kernel size broadens the influence to a wider region, capturing more context but diluting the weight concentration on the center.

In our design, the kernel size corresponds to \(\sigma\), creating an intrinsic relationship between the two parameters. This coupling means the effects of \(\sigma\) and kernel size cannot be isolated, making it challenging to intuitively infer the specific impact of each on the final performance. We evaluate the effect by different \(\sigma\) setting.\begin{table}[h]
    \centering
    \resizebox{0.45\textwidth}{!}{ 
    \renewcommand{\arraystretch}{0.9}
    \setlength{\tabcolsep}{6pt}
    \begin{tabular}{c|c|c|c|c}
        \toprule
        \textbf{Method} & \textbf{Kernel Size} & \textbf{IoU $\uparrow$} & \textbf{IoU\textquotesingle$\uparrow$}
        & \textbf{FLOPs $\downarrow$} \\
        \midrule
        \multirow{4}{*}{DeepLab} 
        & 17 & 0.48 & 0.49 & 2.38M \\
        & 25 & 0.50 & 0.51 & 5.12M \\
        & \cellcolor{green!10} 33 & \cellcolor{green!10} 0.52 & \cellcolor{green!10} 0.53 & 8.92M \\
        & 41 & 0.52 & 0.54 & 13.77M \\
        \midrule
        \multirow{4}{*}{PSPNet} 
        & 17 & 0.45 & 0.45 & 2.38M \\
        & 25 & 0.49 & 0.51 & 5.12M \\
        & \cellcolor{green!10} 33 & \cellcolor{green!10} 0.49 & \cellcolor{green!10} 0.50 & 8.92M \\
        & 41 & 0.48 & 0.48 & 13.77M \\
        \midrule
        \multirow{4}{*}{HRNet} 
        & 17 & 0.45 & 0.46 & 2.38M \\
        & 25 & 0.47 & 0.47 & 5.12M \\
        & \cellcolor{green!10} 33 & \cellcolor{green!10} 0.47 & \cellcolor{green!10} 0.49 & 8.92M \\
        & 41 & 0.46 & 0.49 & 13.77M \\
        \midrule
        \multirow{4}{*}{SegFormer-B4} 
        & 17 & 0.41 & 0.44 & 2.38M \\
        & 25 & 0.43 & 0.47 & 5.12M \\
        & \cellcolor{green!10} 33 & \cellcolor{green!10} 0.46 & \cellcolor{green!10} 0.48 & 8.92M \\
        & 41 & 0.45 & 0.48 & 13.77M \\
        \midrule
        \multirow{4}{*}{SegFormer-B5} 
        & 17 & 0.47 & 0.47 & 2.38M \\
        & 25 & 0.50 & 0.52 & 5.12M \\
        & \cellcolor{green!10} 33 & \cellcolor{green!10} 0.51 & \cellcolor{green!10} 0.52 & 8.92M \\
        & 41 & 0.49 & 0.50 & 13.77M \\
        \bottomrule
    \end{tabular}
    }
    \caption{Effect of Gaussian kernel size deployed by the sampler on FSNet. FLOPs are estimated for the kernel-involved operation on a single image of size $64 \times 128$.}

    \label{tab:kernel_size}
\end{table}

\subsection{Ablation Study on Gaussian Kernel}
We evaluate the effect of increasing the size of the Gaussian kernel on performance and sampling efficiency, while keeping the standard deviation \(\sigma\) fixed at \(\frac{1}{2}\) of the kernel size. Table~\ref{tab:kernel_size} summarizes the results, demonstrating the relationship between performance and sampling efficiency on the CityScapes dataset. As shown, as the kernel size increases, the sampling efficiency decreases quadratically, with a \(5.79\times\) increase in computational cost observed when the kernel size grows from 17 to 41. Furthermore, DeepLab achieves optimal performance at a kernel size of 41, PSPNet at 25, while other models achieve optimal performance at 33. Despite these differences, the performance across models is relatively similar in both IoU and IoU\textquotesingle.

\section{More Visulization Result of FSNet}
\begin{figure*}[h]
\centering
\includegraphics[width=1\linewidth]{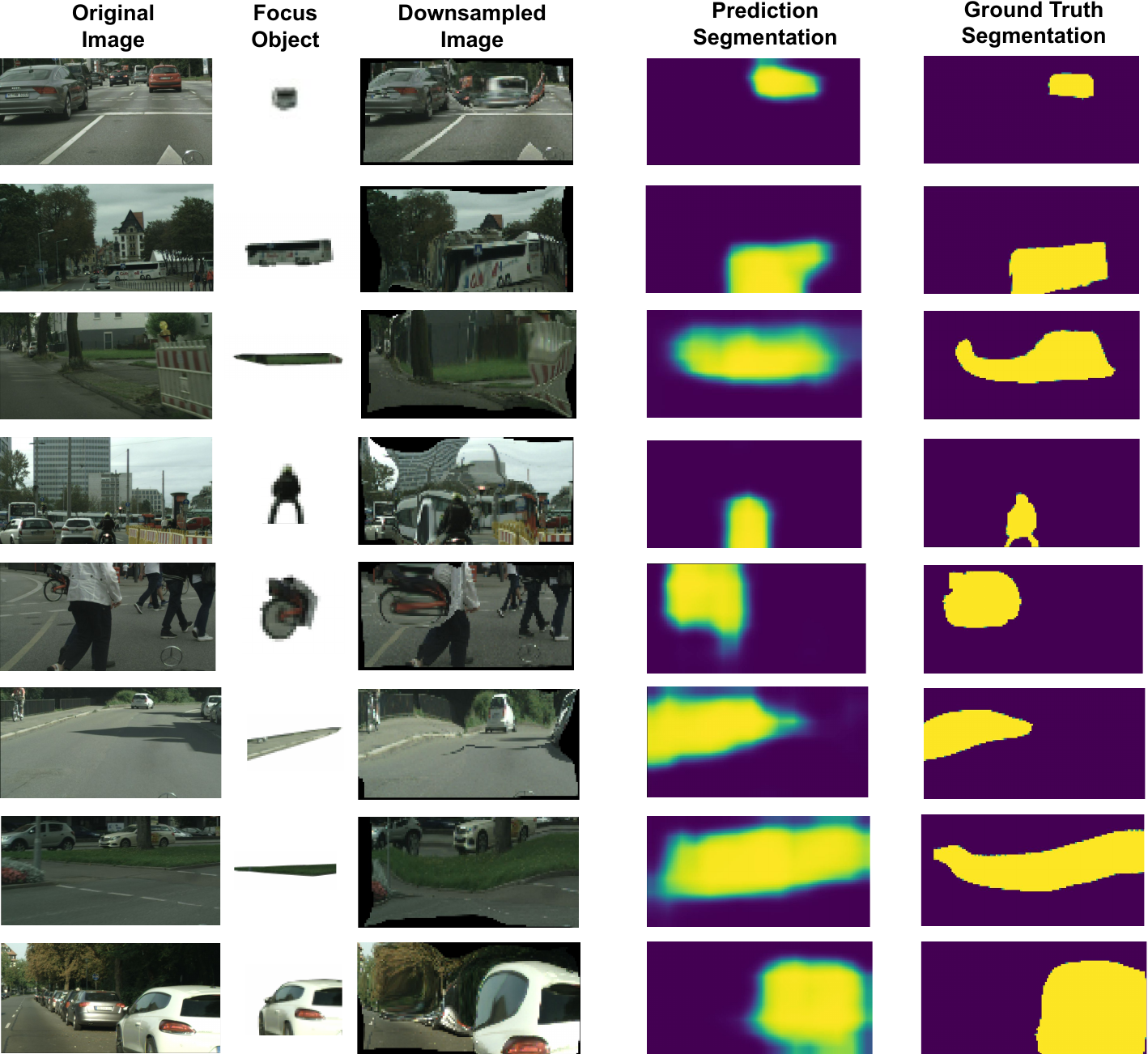}
\caption{Qualitative results of FSNet on CityScapes with different focus objects. The focus objects from top to bottom are: Car, Bus, Traffic Light, Grass, Rider, Bike, Car, Sidewalk, Grass, and Car.}
\label{fig:city_visual}
\end{figure*}

\begin{figure*}[h]
\centering
\includegraphics[width=0.95\linewidth]{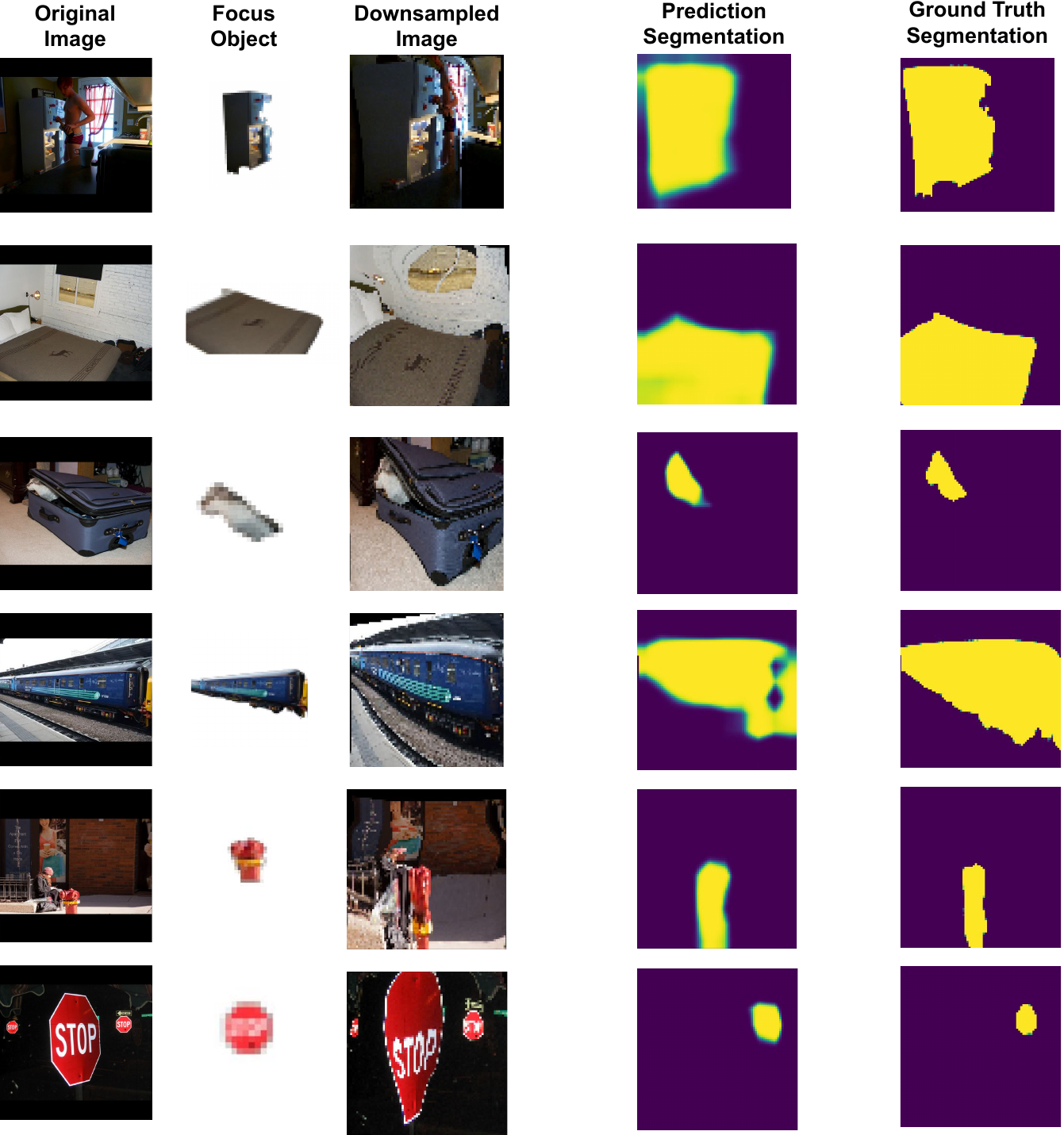}
\caption{Qualitative results of FSNet on LVIS with different focus objects. The focus objects from top to bottom are: Refrigerator, Refrigerator, Mattress, Cat, Train, Fire Hydrant and Traffic Sign.}
\label{fig:lvis_visual}
\end{figure*}

In Figures~\ref{fig:city_visual} and~\ref{fig:lvis_visual}, we present representative examples of original images, their downsampled counterparts, and the corresponding segmentation results produced by FSNet with the DeepLab backbone. 

\clearpage

{
    \small
    \bibliographystyle{ieeenat_fullname}
    \bibliography{main}
}
